\documentclass[conference]{IEEEtran}
\IEEEoverridecommandlockouts
\usepackage{cite}
\usepackage{amsmath,amssymb,amsfonts}
\allowdisplaybreaks
\usepackage{algorithmic}
\usepackage{graphicx}
\usepackage[caption=false,font=footnotesize]{subfig}
\usepackage{textcomp}
\usepackage{xcolor}
\usepackage[normalem]{ulem}
\usepackage{color,soul}
\usepackage[hidelinks]{hyperref}
\def\BibTeX{{\rm B\kern-.05em{\sc i\kern-.025em b}\kern-.08em
    T\kern-.1667em\lower.7ex\hbox{E}\kern-.125emX}}
\begin{document}

\title{LMPath: Language-Mediated Priors and \\
Path Generation for Aerial Exploration}

\author{\IEEEauthorblockN{Jonathan A. Diller, Fernando Cladera, Camillo J. Taylor, Vijay Kumar}
\thanks{\hspace*{-\parindent}This work was funded in part by the ARL DCIST CRA W911NF-17-2-0181. All authors are with the GRASP Laboratory, University of Pennsylvania. {\tt\small \{diller, fclad, cjtaylor, kumar\}@seas.upenn.edu}}}

\maketitle

\begin{abstract}
	Traditional autonomous UAV search missions rely on geometric coverage patterns that ignore the semantic context of the target, leading to significant time waste in large-scale environments. In this paper we present LMPath, a pipeline for generating language-mediated exploration priors for Unmanned Aerial Vehicle (UAV) search missions that leverages semantics. Given a basic geofence and an object of interest prompt, LMPath uses generative language models to determine what regions of the environment should contain that object and a foundation vision model ran over satellite imagery to segment sub-regions that form the exploration prior. This prior can then be used to generate UAV paths with various objectives, such as minimizing the expected time to locate the object of interest, maximizing the probability that the object is found given a limited travel distance, or narrowing down the search space to sub-regions that are most likely to contain the object. To demonstrate it's capabilities, we used LMPath to generate various UAV paths and ran them using a real UAV over large-scale environments. We also ran simulations to demonstrate how paths generated using LMPath outperform traditional path planning approaches for search missions.
	The code of LMPath is available open source\footnote{\url{https://github.com/KumarRobotics/LMPath}}.
\end{abstract}


\section{Introduction}
Unmanned aerial vehicle (UAV) mission planning of overhead flight is usually performed manually. The traditional approach to path planning for UAV-search missions is for the human operator to manually select where the vehicle should go using GPS coordinates~\cite{Diller2025Energy,Cladera2025Air}. These conventional methods severely limit the scalability and adaptability of autonomous UAV operations, as manual planning needs to be performed to change the UAV task.

The advent of foundation models that can understand semantics in language and perform zero-shot vision tasks provides a means for rethinking how the user tasks UAVs. Large Language Models (LLMs) can understand the semantics of objects and the environments that they are customarily found in. For example, we expect to find cars in parking lots and along streets. Furthermore, foundation models for image segmentation, such as SAM 3\footnote{\url{https://ai.meta.com/research/sam3}}, have demonstrated the ability to zero-shot segmentation tasks over varying domains. These powerful models, combined with high-resolution satellite imagery, could be used to generate language-mediated exploration priors that can be leveraged for generating UAV paths for search missions.

\begin{figure}[t]
    \centering
    \includegraphics[width=.75\linewidth]{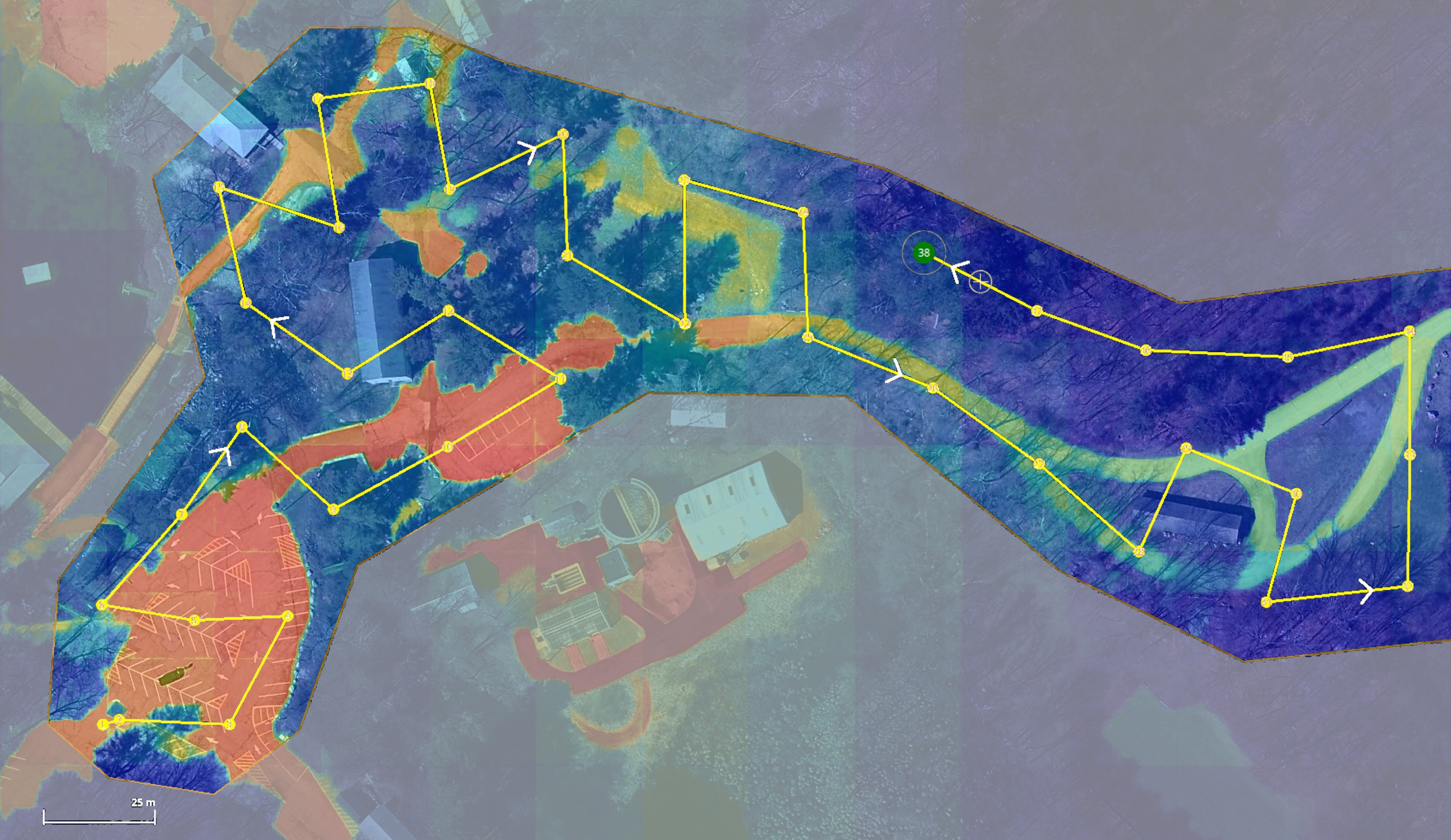}
    \vspace{1em}

    \includegraphics[width=.75\linewidth]{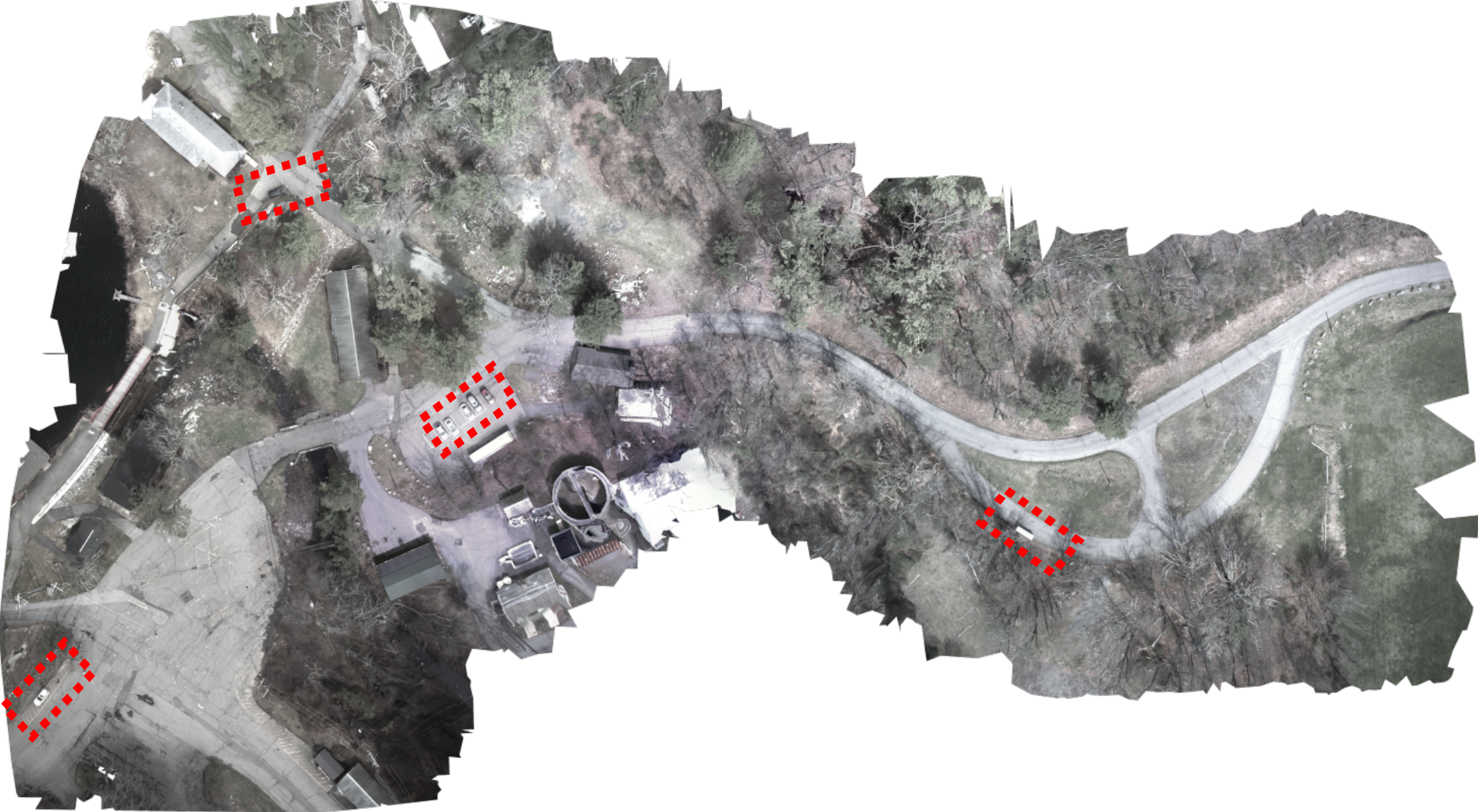}
    \caption{\uline{Top}: search prior generated by LMPath, targeting the label \emph{car}, and UAV path  to minimize expected search time.
    \uline{Bottom}: orthomosaic generated after flying the mission, showing detected cars
    in red.}
\end{figure}

\begin{figure*}[t]
    \centering
    \includegraphics[width=0.9\textwidth]{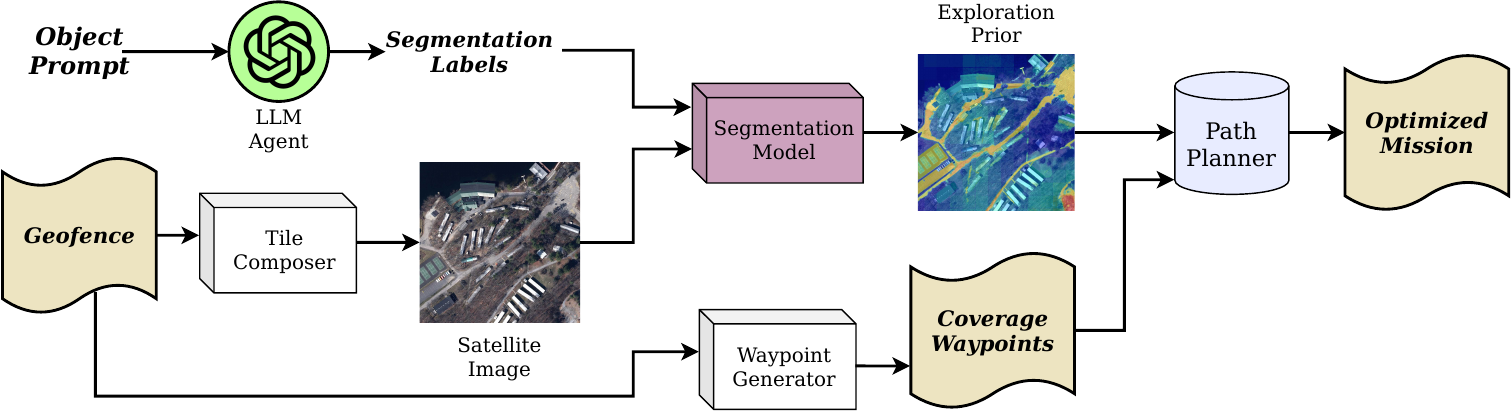}
    \caption{The LMPath pipeline. LMPath takes a user-provided object prompt and geofence bounds for exploration and generates a heatmap that serves as an exploration prior. The exploration prior is then crossed with waypoints that provide sensor coverage of the area of interest and fed into a path planner.}
    \label{fig:pipeline}
\end{figure*}

Existing methods for using foundation models on satellite imagery for UAVs tend to focus on locating larger, static objects that are visible in satellite imagery, such as large buildings~\cite{Sautenkov2025VLPA,Li2025IB}. These methods lack the ability to reason about the semantics between dynamic objects and their environment, such as searching for cars in parking lots. While local reactive navigation methods that utilize language embeddings to guide frontier exploration have shown promise in leveraging semantics~\cite{Taohan2025HALO}, they lack a mechanism to translate high-level environmental common sense into global, optimized search priors for large-scale areas.



In this paper we present LMPath, a pipeline for generating language-mediated exploration priors that leverages semantic-spatial reasoning for optimized UAV path generation. Our contributions are as follows:
\begin{enumerate}
	\item \textbf{A semantic-spatial reasoning framework:} We leverage an LLM to infer semantically relevant labels that provide a reasoning layer between a search target and its likely environmental context.
	\item \textbf{A multi-objective path-finding system:} We integrate these language-mediated priors into an Integer-Linear Program (ILP) to optimize paths for diverse mission objectives, such as minimizing search time.
\end{enumerate}
We demonstrate the system's utility through real-world field tests and high-fidelity simulations where LMPath outperformed semantic-agnostic baselines by up to 88.0\%.

\section{Language-Mediated Priors Framework}
The LMPath pipeline (shown in figure~\ref{fig:pipeline}) consists of an LLM agent for generating semantically relevant labels, a vision foundation model for segmenting regions of the environment that match the semantic labels, a waypoint generator based on sensor coverage, and an ILP for path finding.

\subsection{Generating An Exploration Prior}
The LMPath pipeline initializes by taking in a user-provided target label and geospatial metadata that defines the operational geofence boundaries, no-fly zones, and the UAV's starting location, $b \in \mathbb{R}^2$. This geospatial metadata is formatted as a QGroundControl (QGC) \texttt{plan} mission file. Using these boundaries, a \textit{Tile Composer} queries a web-based map tile service to retrieve and stitch together high-resolution tiles, forming a complete global satellite image of the search area, denoted as $\mathcal{I}$. To generate a heatmap prior for an arbitrary, user-defined target object $\mathcal{O}$ (e.g., \textit{car}), the framework prompts an LLM to infer likely spatial contexts. The LLM outputs a set of semantically relevant labels $L = \{l_1, l_2, \dots, l_k\}$ (e.g., \textit{parking lot}, \textit{road}, \textit{driveway}). This process effectively bridges the gap between the specific search target and the broader semantic features recognizable in overhead imagery.

With the semantic labels $L$ defined, the pipeline employs a foundation segmentation model, SAM 3, to generate a probabilistic heat map over $\mathcal{I}$. Because satellite imagery is typically too large to process in a single forward pass, we implement a sliding window approach. 

Let $\mathcal{W}$ be the set of overlapping image windows extracted from $\mathcal{I}$. For each window $w \in \mathcal{W}$ and each label $l \in L$, SAM 3 generates a binary segmentation mask $M_{w, l}(x,y) \in \{0, 1\}$. To mitigate boundary artifacts and ensure smooth transitions, the overlapping masks are averaged. The aggregated mask for a given label $l$ across the entire image is computed as 
$$M_l(x,y) = \frac{\sum_{w \in \mathcal{W}(x,y)} M_{w,l}(x,y)}{|\mathcal{W}(x,y)|},$$
\noindent
where $\mathcal{W}(x,y)$ represents the subset of windows that contain the pixel coordinate $(x,y)$. The final semantic prior, or heat map, $H(x,y)$, is generated by summing the masks for all labels and normalizing the distribution:
$$H(x,y) = \frac{\sum_{l \in L} M_l(x,y)}{\iint_{\mathcal{I}} \sum_{l \in L} M_l(u,v) \, du \, dv}.$$
\noindent
This results in a 2D probability density function where higher values indicate a greater semantic likelihood of the object $\mathcal{O}$ being present.

\begin{figure*}[!ht]
	\centering
	\subfloat[Geofence \& Launch Position]{
		\includegraphics[width=0.23\textwidth]{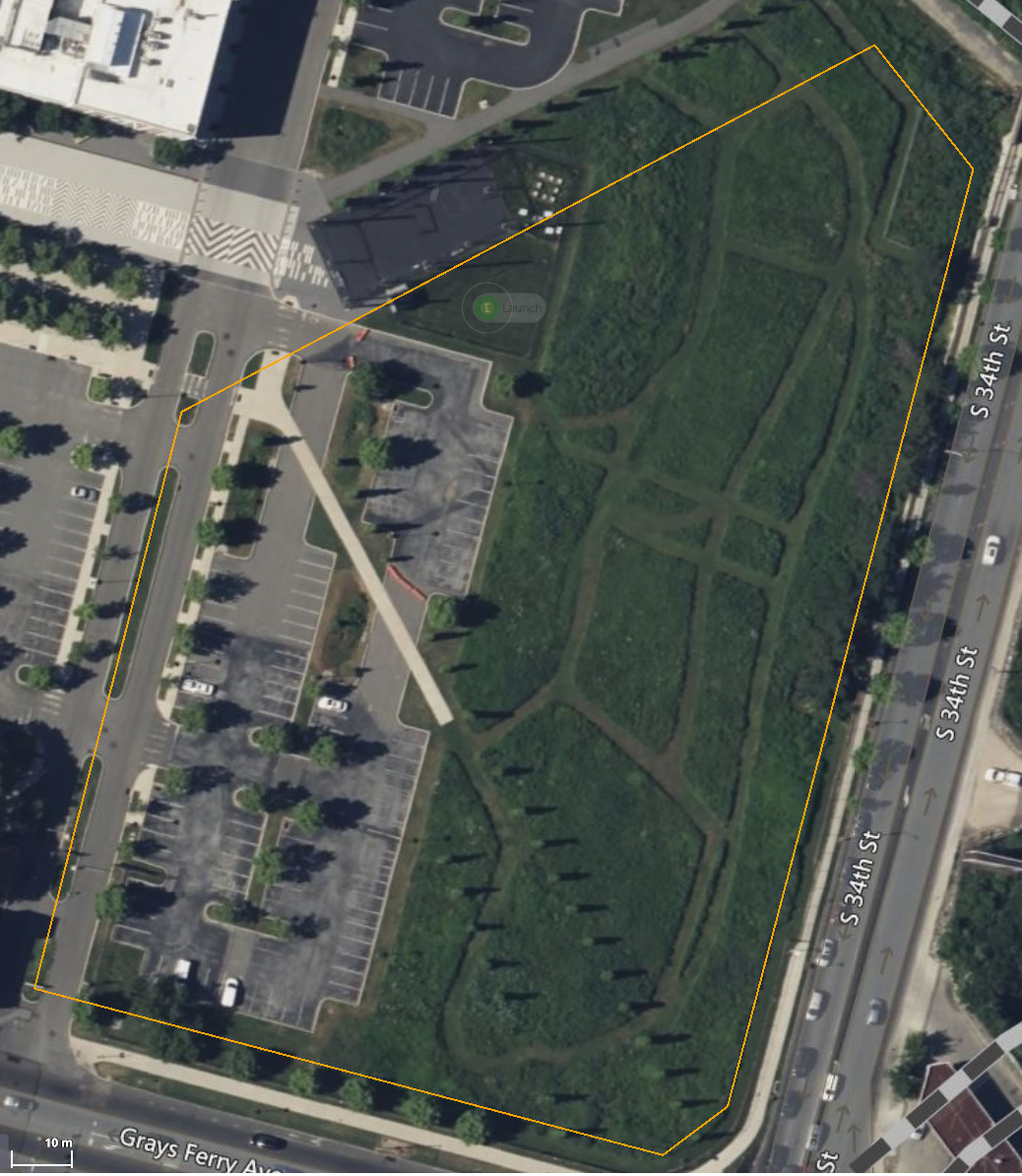}
		\label{fig:penn_fence}
	}
	\hfil 
	\subfloat[Satellite Image]{
		\includegraphics[width=0.225\textwidth]{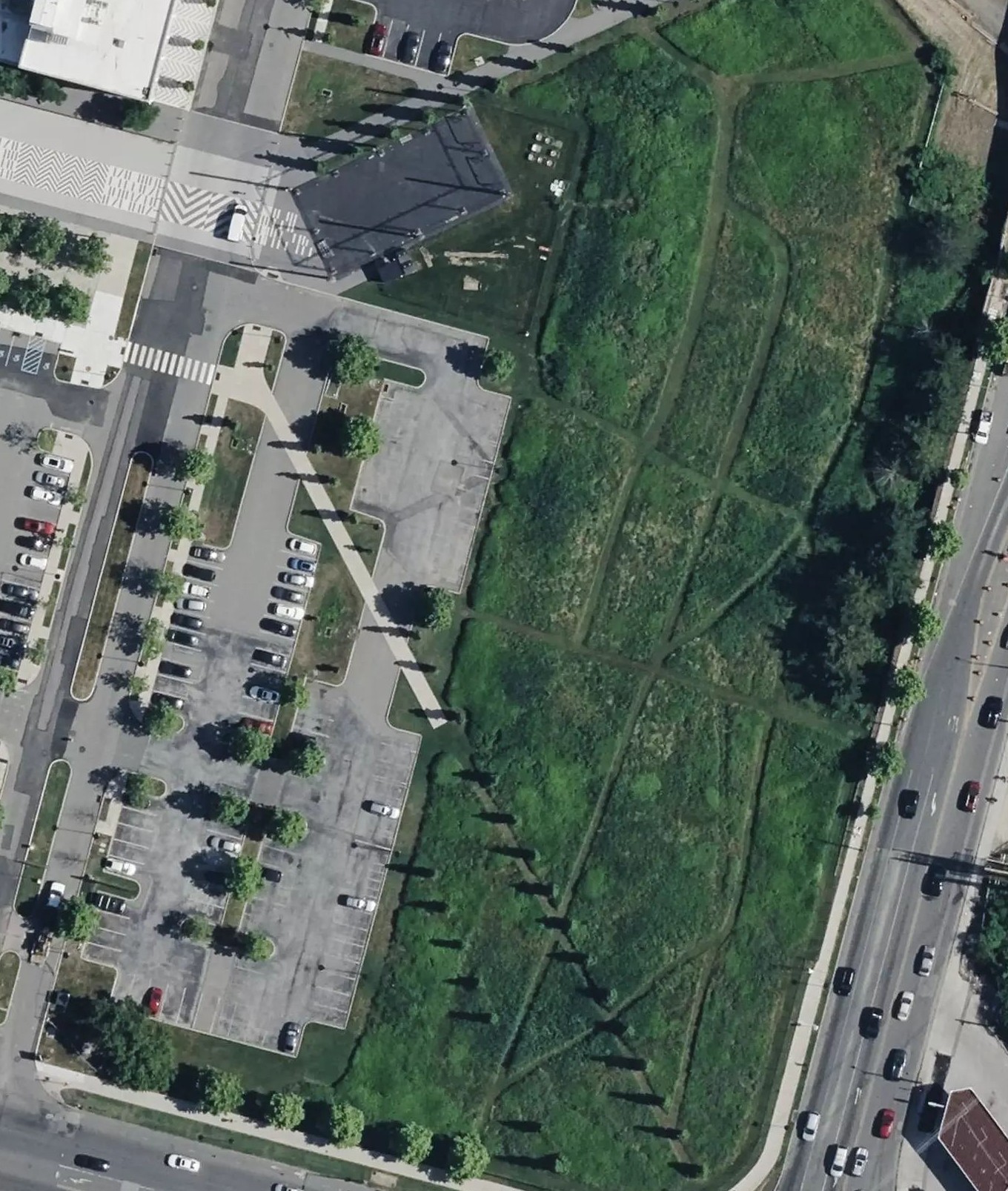}
		\label{fig:penn_tiles}
	}
	\hfil
	\subfloat[LMPath-Generated Heatmap]{
		\includegraphics[width=0.225\textwidth]{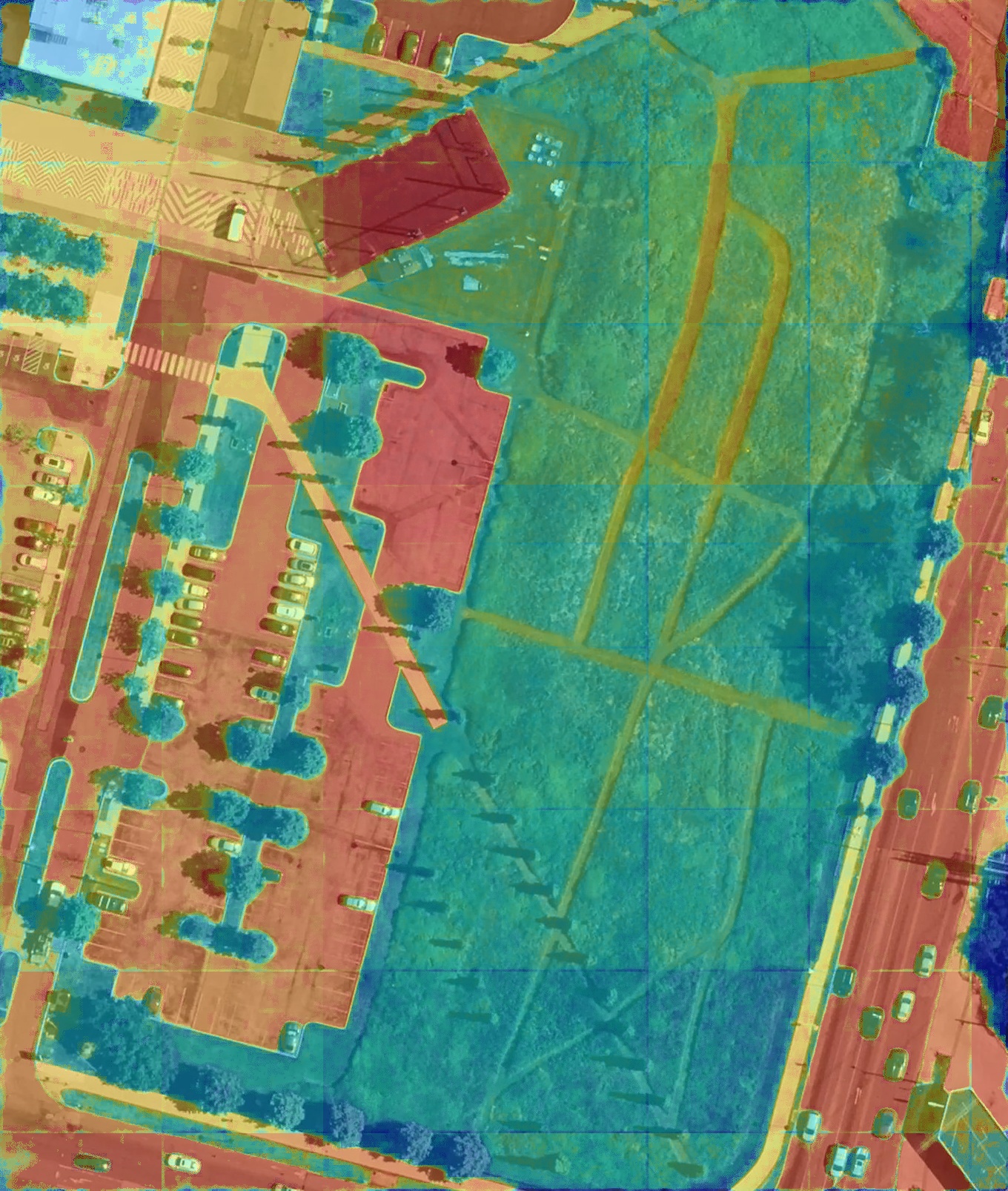}
		\label{fig:penn_mask}
	}
	\hfil
	\subfloat[LMPath-Generated Path]{
		\includegraphics[width=0.23\textwidth]{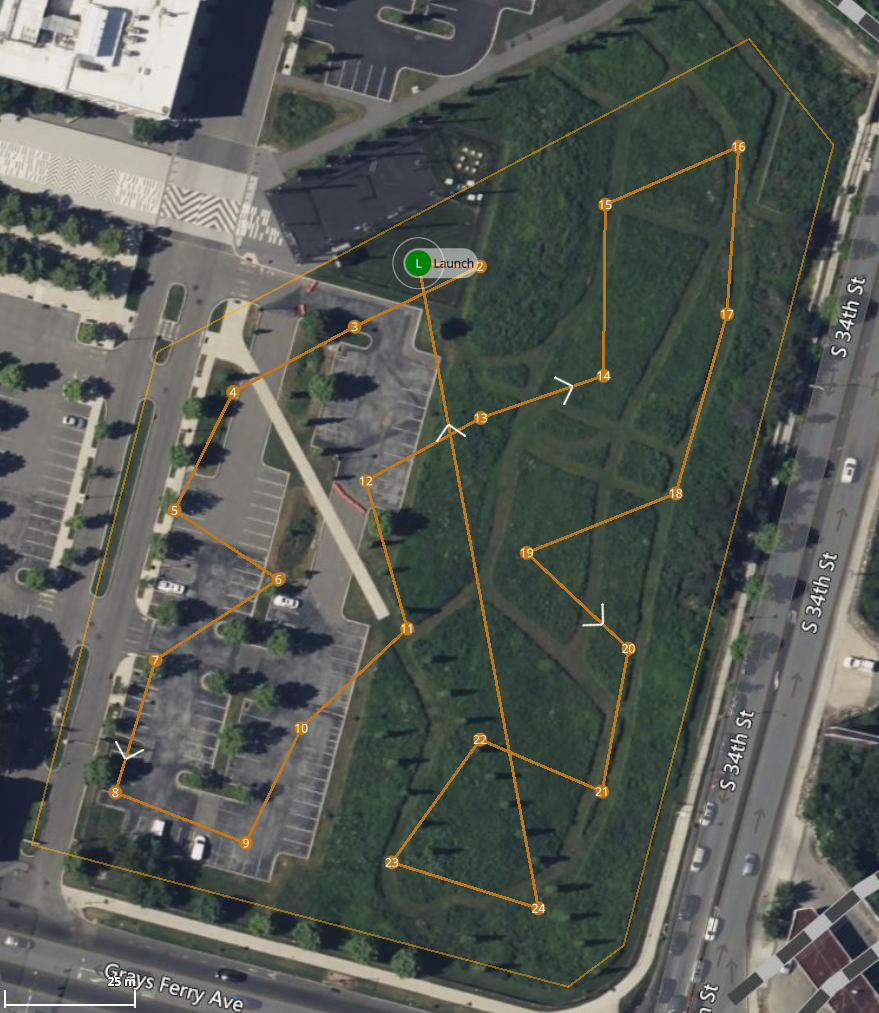}
		\label{fig:penn_path}
	}

	\subfloat[Geofence \& Launch Position]{
		\includegraphics[width=0.23\textwidth]{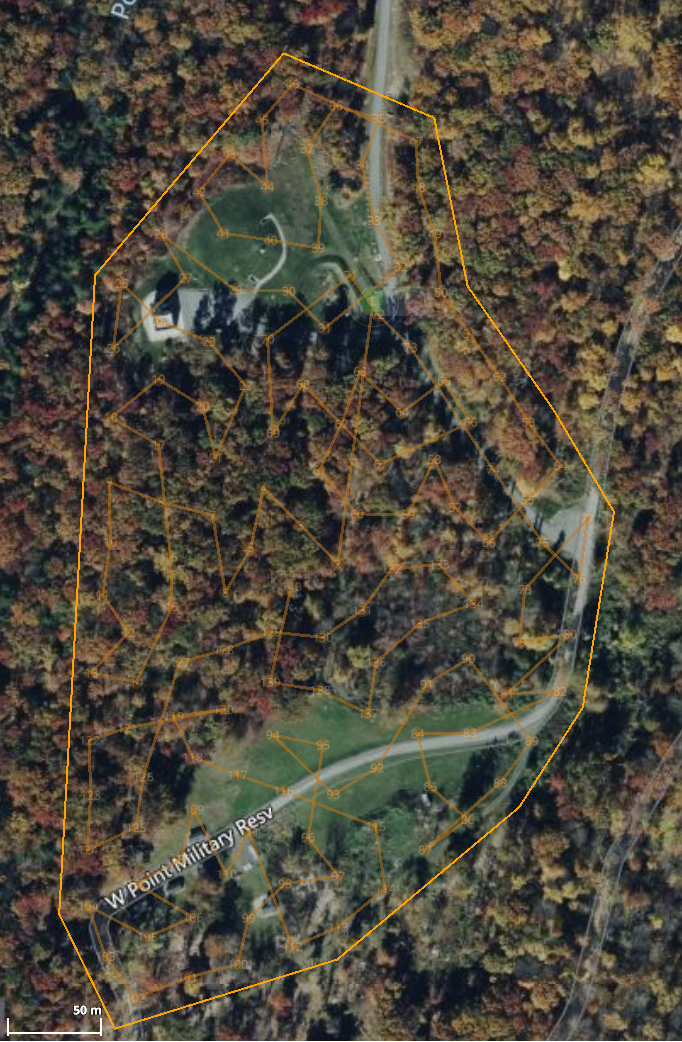}
		\label{fig:range_fence}
	}
	\hfil 
	\subfloat[Satellite Image]{
		\includegraphics[width=0.2\textwidth]{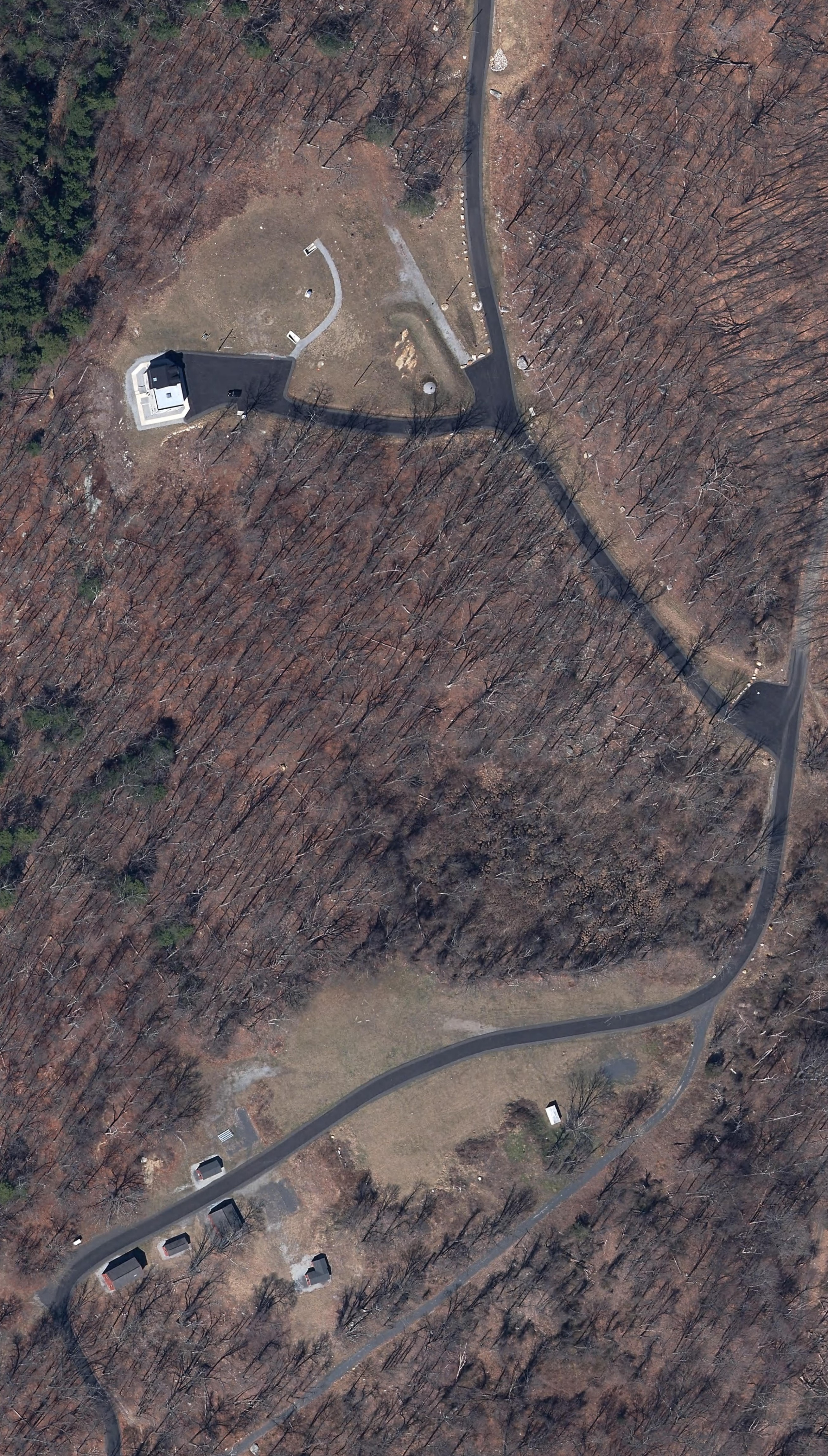}
		\label{fig:range_tiles}
	}
	\hfil
	\subfloat[LMPath-Generated Heatmap]{
		\includegraphics[width=0.2\textwidth]{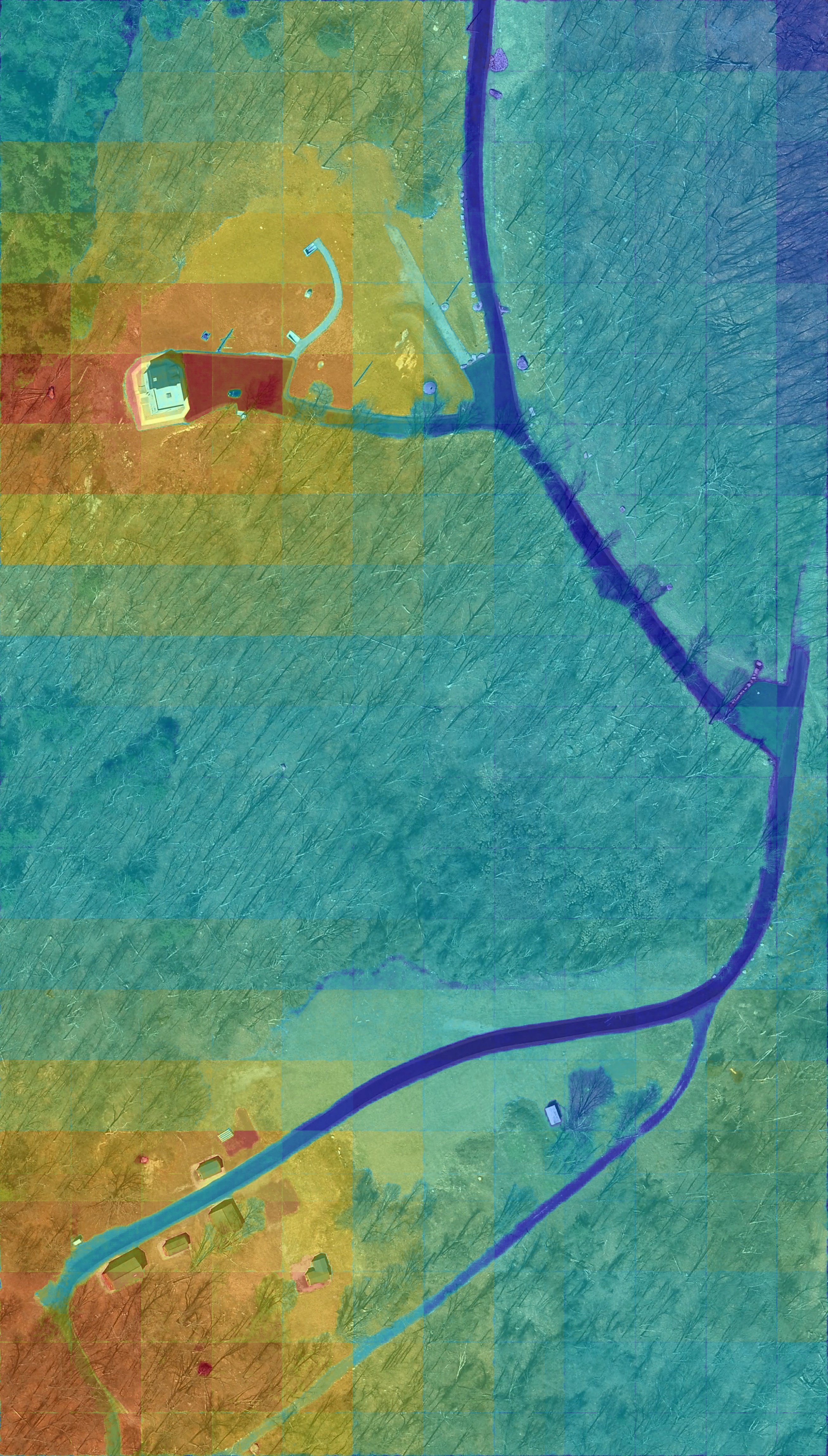}
		\label{fig:range_mask}
	}
	\hfil
	\subfloat[LMPath-Generated Path]{
		\includegraphics[width=0.23\textwidth]{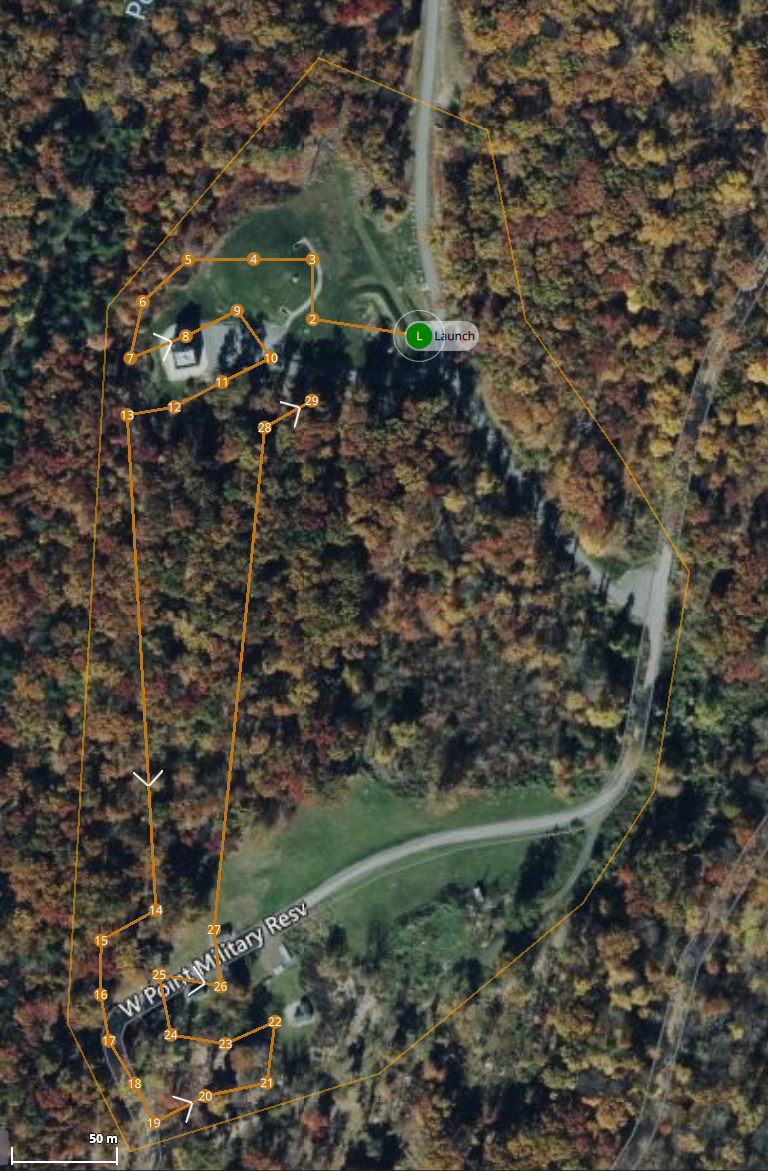}
		\label{fig:range_path}
	}
	\caption{LMPath examples for generating heatmaps and flight paths to find the object ``\textit{car}'' in minimum time (\textit{a}-\textit{d}), and a threshold-reduced flight path to find the object ``\textit{building}.'' (\textit{e}-\textit{h}). The \textit{car} example shows complete coverage but explores parking lots before the grass field, while the \textit{building} example demonstrates the UAV exploring two distinctive sub-regions.}
	\label{fig:range}
\end{figure*}

\subsection{Waypoint Generation and Path Finding}
To translate the continuous heat map $H(x,y)$ into a discrete graph suitable for path planning, a \textit{Waypoint Generator} creates a set of candidate waypoints $N = \{n_1, n_2, \dots, n_m\}$ over the search area. The \textit{Waypoint Generator} sets the number of waypoints based on the UAV's sensor coverage footprint and the total area of $\mathcal{I}$. 

The pipeline then partitions the operational space using a Voronoi tessellation based on $N$. The Voronoi cell $v_i$ corresponding to waypoint $n_i$ is defined as
$$v_i = \{ p \in \mathbb{R}^2 \mid \|p - n_i\| \leq \|p - n_j\| \; \forall j \neq i \}.$$
\noindent
The ``heat'' or probability mass associated with each waypoint, denoted as $p_i \in P$, is calculated by integrating the semantic prior over its corresponding Voronoi cell and normalizing by the cell's area $A(v_i)$ to prevent bias toward disproportionately large cells:
$$p_i = \frac{1}{A(v_i)} \iint_{v_i} H(x,y) \, dx \, dy.$$

The final component of the pipeline is a \textit{Path Planner} for generating UAV paths based on various objectives over \(P\).

\subsubsection{Minimizing Expected Search Time} \label{subsec:minT}
In the context of searching for a specific object of interest, the user may want to minimize the expected time required to locate that object. Let $x_{ij} \in \{0, 1\}$ be a binary decision variable that equals $1$ if the UAV travels from waypoint $n_i$ to $n_j$, and $0$ otherwise. Let $t_{i}$ denote the time that the UAV visits waypoint \(i\). This problem can be formulated as the following ILP:
\begin{equation}
	\min \sum_{i \in N} p_i t_i
\end{equation}
\textit{subject to}
\begin{subequations}
\label{eq:grouped_equations}
\begin{align}
	&\sum_{j \in N} x_{ji} + x_{b,i} = 1, &\forall i \in N \label{cnst:1}\\
	&\sum_{j \in N} x_{ij} + x_{i,b} = 1, &\forall i \in N \label{cnst:2} \\
	&\sum_{i \in N} x_{b,i} = 1 \label{cnst:3} \\
	&x_{ii} = 0, &\forall i \in N \label{cnst:4} \\
	&t_j \geq t_i + \frac{d_{ij}}{v} - \mathcal{M}(1 - x_{ij}), &\forall i, j \in N \label{cnst:5} \\
	&t_i \geq \frac{d_{b,i}}{v} - \mathcal{M}(1 - x_{b,i}), &\forall i \in N \label{cnst:6}
\end{align}
\end{subequations}

Constraints~\ref{cnst:1} and~\ref{cnst:2} specify that waypoint $i$ must be entered and exited, respectively, exactly once, either from another waypoint $j$ or from the base station \(b\). Constraint~\ref{cnst:3} ensures that the UAV goes to exactly one node after leaving the base station while constraint~\ref{cnst:4} prevents single waypoint cycles. If the UAV travels from $i$ to $j$ (i.e., $x_{ij} = 1$), then constraint~\ref{cnst:5} forces $t_j$ to be greater than or equal to $t_i$ plus the time to travel distance \(d_{ij}\) at speed \(v\) using an arbitrarily large value \(\mathcal{M}\). Constraint~\ref{cnst:6} enforces the same restriction on the first waypoint in the route.

\subsubsection{Target-Focused Path in Large Environments} \label{subsec:maxTarget}
LMPath can generate missions for large-scale environments where covering the entire space is not expected to be as helpful as focusing on mission-relevant sub-regions. To generate an efficient, mission-focused paths in large environments the \textit{Path Planner} filters out waypoints \(i \in N\) where \(p_i < \rho\), for some threshold value \(\rho\). The remaining waypoints can then be treated as an input to the Traveling Salesman Problem (TSP), for which there are many well studied algorithms.

\section{Experimental Evaluation}

\subsection{Experimental Setup}
We utilized SAM 3 as our foundation segmentation model and GPT-4o-mini as the LLM agent. The path planner was implemented in C++ using the Gurobi solver (version 13.01). For the geospatial raster tile service, Mapbox provided high-quality satellite tiles at high zoom levels, though ESRI World Imagery and Google Maps were also evaluated. Empirically, a 75\% overlap in the sliding window mask mosaic and a sub-region size of 100 square meters yielded consistent heat map priors across various environments.

\subsection{Real-World Validation}
To evaluate the LMPath pipeline, we generated several path plans for distinct operational missions and ran them on a UAV. Figures~\ref{fig:penn_fence} through~\ref{fig:penn_path} show an example of searching for a car by solving the ILP presented in Section~\ref{subsec:minT}. The UAV path initially goes over the parking lot before covering the remainder of the search space, which mostly consistent of a grass field. Figures~\ref{fig:range_fence} through~\ref{fig:range_path} show an example of searching for buildings within a large-scale environment (300 m \(\times\) 450 m) using the path planning approach described in Section~\ref{subsec:maxTarget}. The UAV path covers two large, open spaces and ignores large sections of forest. These missions were successfully executed in real-world field tests using a Falcon 4 UAV with a nadir-viewing RGB camera~\cite{cladera2025evmapper}.

\subsection{Simulation Environments}
\begin{figure}[!t]
    \centering
    \subfloat[Exploration prior]{
        \includegraphics[width=0.45\columnwidth]{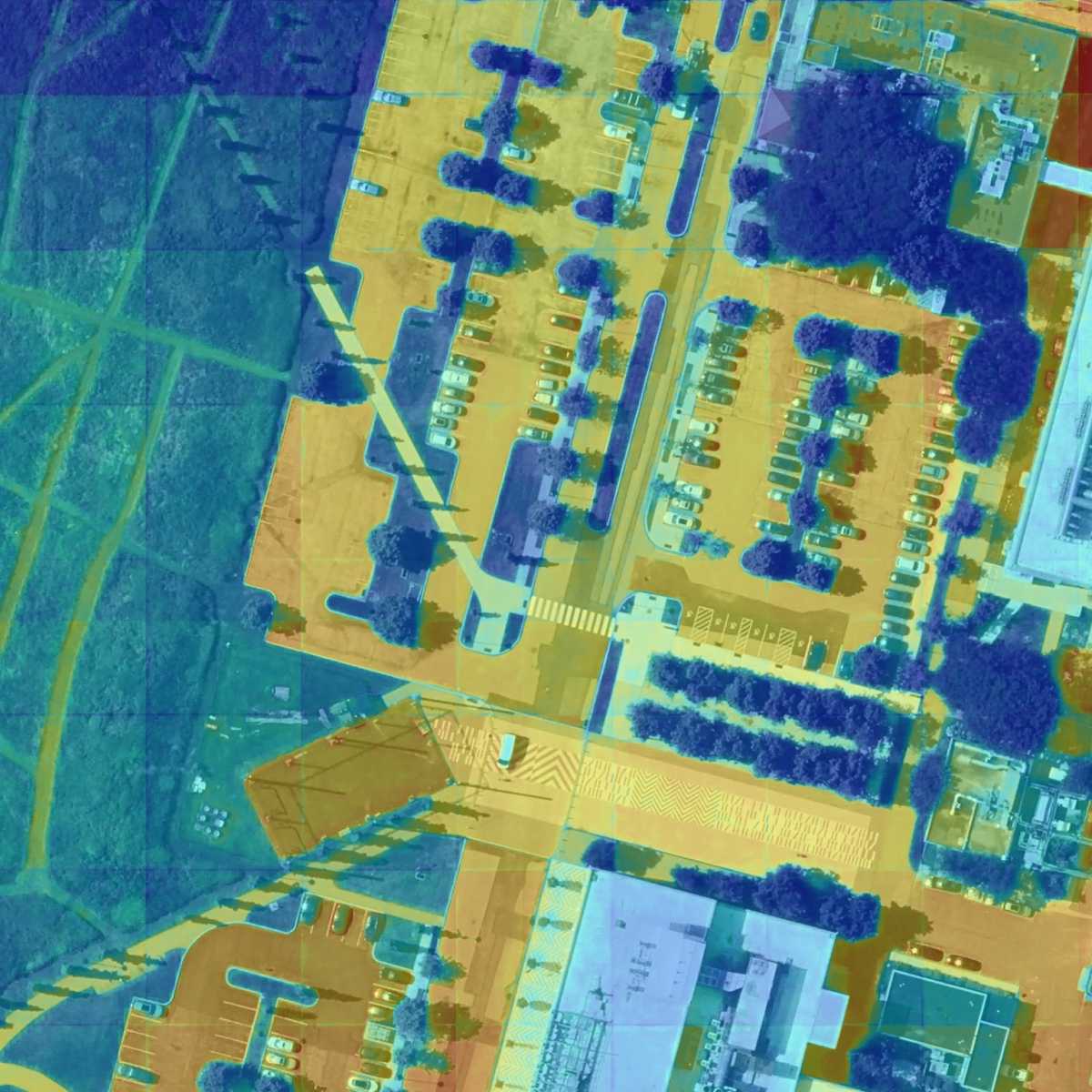}
        \label{fig:col_img1}
    }
    \hfil
    \subfloat[Drone simulator]{
        \includegraphics[width=0.45\columnwidth]{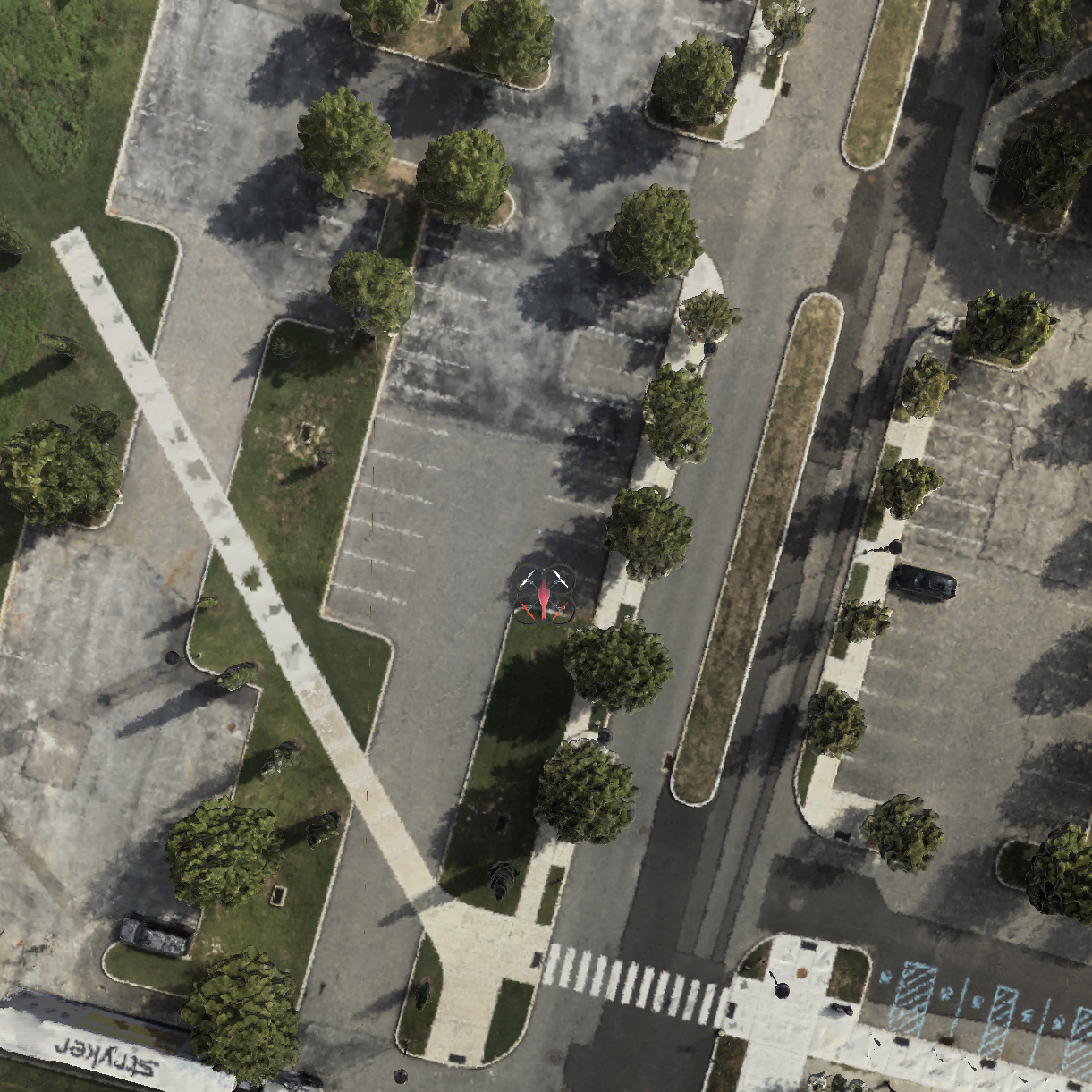}
        \label{fig:col_img2}
    }
    \caption{Exploration prior from real satellite image (\textit{a}) and drone simulator for \textit{Industrial Park}.}
    \label{fig:penn_sim}
\end{figure}

To quantitatively assess the framework's ability to minimize search time, we conducted a series of Gazebo simulations where the UAV was tasked with locating a randomly selected target car. The simulations were evaluated across two distinct environments:
\begin{itemize}
	\item \textit{PolyCity:} A synthetic, low-complexity urban layout.
	\item \textit{Industrial Park:} A high-fidelity 3D mesh reconstructed from actual UAV-captured imagery (Figure~\ref{fig:penn_sim}).
\end{itemize}

Crucially, for the real-world mesh environment, LMPath was provided with standard web-sourced satellite imagery of that same geographic location. This established a highly realistic operational scenario where the generated semantic prior covers the correct spatial area but reflects the environment at a different point in time than the simulated world.

We evaluated the efficiency of the LMPath framework by measuring the total flight time required to successfully locate a randomly selected target car. We repeated this experiment 50 times, using the same UAV path but selecting different cars using a uniform distribution. We compared our approach against a baseline TSP planner, which calculates an efficient route to uniformly cover the entire operational space without the guidance of a semantic prior. The path generated using LMPath found the target faster than the baseline 66.0\% of the time in \textit{PolyCity} and 88.0\% of the time in the \textit{Industrial Park}. The results demonstrate that LMPath can significantly reduce the expected search time by intelligently prioritizing high-probability regions over a semantic-agnostic exhaustive coverage approach.

\section{Conclusions}
In this paper, we introduced LMPath, a novel pipeline that leverages Large Language Models and foundation vision models to generate semantic exploration priors for UAV search missions. By linking user-defined prompts with satellite imagery, LMPath transforms continuous environments into discrete, probability-weighted graphs for advanced path planning. Both our real-world Falcon 4 field tests and high-fidelity simulations demonstrate that intelligently targeting high-probability sub-regions significantly outperforms traditional, non-semantic coverage methods.

One of the limitations of LMPath is that it can only segment locations that are clearly visible from the given satellite image. This could become problematic in areas where large buildings or vegetation create occlusions. Future work will focus on merging the LMPath exploration prior with local, frontier exploration algorithms~\cite{Taohan2025HALO}. Additionally, the current heat map formulation assumes uniform weighting to labels and masks. Future iterations could incorporate label confidence and conditional dependencies (e.g., favor ``driveway'' when adjacent to ``building'') to further refine the search prior.

\bibliographystyle{IEEEtran}
\bibliography{references}

\end{document}